\pdfoutput=1
\documentclass[runningheads,a4paper]{llncs}
\usepackage{times}
\usepackage{latexsym}
\usepackage{url}

\usepackage{graphicx}
\graphicspath{{figs/}}

\usepackage{url}
\usepackage{enumitem}

\usepackage[T2A,T1]{fontenc}
\usepackage[utf8x]{inputenc}
\usepackage[russian,english]{babel}

\title{Stance Prediction for Russian: Data and Analysis}

\authorrunning{N. Lozhnikov, L.Derczynski, M.Mazzara}

\author{
Nikita Lozhnikov \inst{1},
Leon Derczynski \inst{2} and
Manuel Mazzara\inst{1}
\institute{Innopolis University, Russian Federation
 \\
\{n.lozhnikov, m.mazzara\}@innopolis.ru
\and ITU Copenhagen, Denmark\\
leod@itu.dk}}

\date{}

\begin{document}
\maketitle

%!TEX root = coling2018.tex

\begin{abstract}
Stance detection is a critical component of rumour and fake news identification. 
It involves the extraction of the stance a particular author takes related to a given claim, both expressed in text. 
This paper investigates stance classification for Russian. 
It introduces a new dataset, RuStance, of Russian tweets and news comments from multiple sources, covering multiple stories, as well as text classification approaches to stance detection as benchmarks over this data in this language. 
As well as presenting this openly-available dataset, the first of its kind for Russian, the paper presents a baseline for stance prediction in the language.
\end{abstract}

%!TEX root = rustance.tex

\section{Introduction}

The web is rife with half-truths, deception, and lies.
The rapid spread of such information, facilitated and accelerated by social media can have immediate and  serious effects. 
Indeed, such false information affects perception of events which can lead to behavioral manipulation~\cite{forbes}. 
The ability to identify this information is important, especially in the modern context of services and analyses that derive from claims on the web~\cite{2017arXiv170400656Z}.

However, detecting these rumours is difficult for humans, let alone machines. 
Evaluating the veracity of a claim in for example social media conversations requires context --  e.g. prior knowledge -- and strong analytical skills~\cite{mrowcastance}. 
One proxy is ``stance". 
Stance is the kind of reaction that an author has to a claim. 
Measuring the stance of the crowd as they react to a claim on social media or other discussion fora acts as a reasonable proxy of claim veracity.

The problem of stance detection has only been addressed for a limited range of languages: English, Spanish and Catalan \cite{ferreira2016emergent,anta2013sentiment,taule2017overview}. 
With these, there are several datasets and shared tasks. 
While adopting now more mature standards for describing the task and structuring data, RuStance enables stance prediction in a new context.

Debate about media control and veracity has a strong tradition in Russia; the populace can be vocal, for example overthrowing an unsatisfactory ruling empire in 1918.
Veracity can also be questionable, for example in the case of radio transmitters in key areas being used to send targeted messages before the internet became the medium of choice~\cite{enikolopov2011media}.
Indeed, news on events and attitudes is Russia is often the focus of content in US and European media, with no transparent oversight or fact checking.
The context is therefore one that may benefit greatly from, or least be highly engaged with, veracity and stance technology.

This paper relates the construction of a stance dataset for Russian, {\sc RuStance}.
The dataset is available openly for download, and accompanied by baselines for stance prediction on the data and analysis of the results.
%!TEX root = rustance.tex

\section{Data and Resources}
\label{chap:datasource}

Before collecting data, we set the scope of the task and criteria for data likely to be ``interesting".
Collection centred around a hierarchical model of conversation, with stories at the top having source rumours/claims, which are referenced as ``source tweets" in prior Twitter-centric work~\cite{zubiaga2016stance}, that are the root of discussion threads, with responses potentially expressing a stance toward that source claim. The dataset is available on GitHub. \footnote{https://github.com/npenzin/rustance}

\subsection{Requirements}

The data was collected during manual observation in November 2017. 
Tweets that started a useful volume of conversationl activity on a topic we had been observing were considered ``interesting'', and were added to the list of the potentially valuable sources.

For individual messages, we needed to determine the objective support towards a rumour, an entire statement, rather than individual target concepts. 
Moreover, we were to determine additional response types to the rumourous tweet that are relevant to the discourse, such as a request for more information (questioning) and making a comment (C), where the latter doesn’t directly address support or denial towards the rumour, but provides an indication of the conversational context surrounding rumours. 
For example, certain patterns of comments and questions can be indicative of false rumours and others indicative of rumours that turn out to be true.

Following prior work~\cite{qazvinian2011rumor,lukasik2016hawkes}, we define Stance Classification as predicting a class (label) given a text (features) and topic (the rumour).
Classes are \textit{Support}, \textit{Deny}, \textit{Query}, \textit{Comment}.

\begin{itemize}
  \item \textbf{Support}: the author of the response supports the veracity of the rumour, especially with facts, mentions, links, pictures, etc. For instance, ``Yes, that's what BBC said.''
  \item \textbf{Deny}: the author of the response denies the veracity of the rumour, the opposite case of the Support class. For instance, ``Under the bed???? I've been there and there were not any monsters!''
  \item \textbf{Query}: the author of the response asks for additional evidence in relation to the veracity of the rumour. This one is usually is said in a qustionable  manner. For instance, ``Could you provide any proof on that?''
  \item \textbf{Comment}: the author of the response makes their own comment without a clear contribution to assessing the veracity of the rumour. The most common class, but not the worst. The examples of the class usually contains a lot of emotions and personal opinions, for example, ``Hate it. This is awesome!'' - tells us nothing about the veracity.
\end{itemize}

\begin{figure*}
\framebox{%
  \parbox[t]{\textwidth}{%
  \addvspace{0.5cm} \centering
  \textbf{Claim}: The Ministry of Defense published irrefutable evidence of US help for ISIS.
  \begin{itemize}
    \item \textit{Reply 1}: Come'on. This is a screenshot from ARMA. \textbf{[deny]}. 
    \item \textit{Reply 2}: Good job! \textbf{[comment]}
    \item \textit{Reply 3}: Is that for real? \textbf{[query]}
    \item \textit{Reply 4}: That's it! RT also say so \textbf{[support]}
  \end{itemize}    
  }%
}
\caption{A synthetic example of a claim and related reactions (English).}
\end{figure*}

\begin{figure*}
\begin{otherlanguage*}{russian}
\small
\framebox{%
    \parbox[t]{\textwidth}{%
    \addvspace{0.5cm} \centering
    \textbf{Claim}: \#СИРИЯ Минобороны России публикует неоспоримое подтверждение обеспечения Соединенными Штатами прикрытия боеспособны… https://t.co/auTz1EBYX0
    \begin{itemize}
      \item \textit{Reply 1}: Это не фейк, просто перепутали фото. И, кстати, заметили и исправили. Это фото из других ново… https://t.co/YtBxWebenL \textbf{[deny]}. 
      \item \textit{Reply 2}: Министерство фейков, пропаганды и отрицания. \textbf{[comment]}
      \item \textit{Reply 3}: Что за картинки, на которых вообще не понятно - кто,где и когда? \textbf{[query]}
      \item \textit{Reply 4}: Эту новость даже провластная Лента запостила, настолько она ржачная) \textbf{[support]}
    \end{itemize}    
    }%
  }
\end{otherlanguage*}
  
    \caption{A real example of a claim and related reactions (Russian).}
\end{figure*}

\subsection{Sources}

In order to create variety, the data is drawn from multiple sources. This increases the variety in the dataset, thus in turn aiding classifiers in generalizing well. For RuStance, the sources chosen were: Twitter, Meduza~\cite{meduza}, Russia Today [RT]~\cite{russiatoday}, and selected posts that had an active discussion.

\begin{itemize}
	\item \textbf{Twitter}: is one of the most popular sources of claims and reactions. 
	%It has a strict policy on the length of posts which means users have to compact and shrink their thoughts. 
	We paid attention to a well-known ambiguous and false claims, for example the one ~\cite{guardian2017russia} mentioned -- Russian Ministry Of Defense.
  \item \textbf{Meduza}: is an online and independent media located in Latvia and focused on Russian speaking audience. 
  Meduza is known to be in opposition to Kremlin and Russian politicians. 
  The media has ability for registered users to leave comments to a particular events. 
  We collected comments on some popular political events that were discussed more than the others ~\cite{meduza2017mod,meduza2017mod2}.
  \item \textbf{Russia Today}: is an international media that targets world wide auditory. 
  It is supposed that the editors policy of Russia Today is to support Russian goverment. 
  Its main topics are politics and international relation which means there are always debates. 
  We gathered some political and provocative publications with a lot of comments~\cite{rt2017}.
\end{itemize}

% \begin{table}[H]
%   \centering
%   \caption{Data Source distribution}
%   \label{data-source-distribution-table}
%   \begin{tabular}{ccc}
%     \textbf{Russia Today} & \textbf{Meduza} & \textbf{Twitter} \\ \hline
%     15\%                  & 45\%            & 40\%            
%   \end{tabular}
% \end{table}

To capture claims and reactions in Twitter we used software developed as part of the PHEME project~\cite{derczynski2014pheme} which allows to download all of the threads and replies of a claim-tweet. 
For other sources we downloaded it by hand or copied.

The dataset sources are Twitter and Meduza with 700 and 200 entities respectively. 

Firstly, Twitter is presented with over 700 interconnected replies, i.e. replies both to the claim and to other replies. 
The latter might be a cause of a large number of arguments and aggression (i.e. have high emotional tension) and as a result, the replies are poorly structured from the grammatical perspective, contain many non-vocabulary words in comparison with national or web corpora. 
Tweets that were labeled as "support" and "deny" tend to have links to related sources or mentions. 
Users of Twitter also use more multimedia, which brings in auditory content not included in this text corpus.

Secondly, Meduza comments are discovered to be more grammatically correct and less aggressive but still non-neutral and sarcastic.
Meduza users are mostly deanonymized, but unfortunately, this is our empirical observation and not mentioned in the data.
Comments on the articles vary in amounts of aggression, however still less aggressive than tweets. 
We hypothesize that this is caused by the fact that news articles provide more context and have teams of editors behind it. 
Users of Meduza tend to provide fewer links and other kinds of media, which may be due to the user interface of the site, or a factor of the different nature of social interactions on this platform.

Finally, Russia Today content is very noisy, and difficult to parse.
This provided the smallest contribution to the dataset, and had the least structure and coherence in its  commentary.
Overall, the dataset contains both structured and grammatically correct comments and unstructured messy documents.
This is indicative of a good sample; one hopes that a dataset for training tools that operate over social media will contain a lot of the noise characteristic of that text type, enabling models equipped to handle the noise in the wild.

Typical headlines for collection included (translated):

\begin{itemize}[noitemsep,topsep=0pt,parsep=0pt,partopsep=0pt]
    \item The Ministry of Defense accused ``a civil co-worker" in the publication of the screenshot from the game instead of the photo of the terrorists. And presented a new ``indisputable evidence";
    \item The Ministry of Defense posted an ``indisputable" evidence of US cooperation with ISIS: a screenshot from a mobile game;
    \item  The Bell has published a wiretapping of Sechin and Ulyukaev phone calls;
    \item Navalny seized from ``Life" 50 thousand rubles. He didn't receive this money as a ``citizen journalist" for shooting himself;
    \item ``If the commander in chief will call into the last fight, Uncle Vova, we are with you." Deputy of the Duma with the Cadets sang a song about Putin;
    \item ``We are very proud of breaking this law." Representatives of VPN-services told "Medusa" why they are not going to cooperate with Russian authorities;
    \item Muslims of Russia suggested to teach ``The basics of religious cultures" from 4 til 11 grade;
    \item ``Auchan" (supermarket) will stop providing free plastic bags.
\end{itemize}

\begin{table}
  \centering
  \caption{Dataset class distribution}
  \label{my-label}
  \begin{tabular}{llll}
    \textbf{Support} & \textbf{Deny} & \textbf{Query} & \textbf{Comment} \\ \hline
    58 (6\%)         & 46 (5\%)      & 192 (20\%)     & 662 (69\%)
  \end{tabular}
\end{table}

The most valuable classes -- \textbf{Support} and \textbf{Deny} -- are outnumbered by more general-purpose classes. 
This is similar to the class distribution that other  stance classification datasets usually have~\cite{ferreira2016emergent,anta2013sentiment,taule2017overview}. 
Indeed, FNC-I, the Fake News Challenge dataset~\cite{fakenewschallenge}, also has a quite similar class distribution: 70\% - comments, 20\% - queries, 10\% - support \& deny. 

In the interests of describing the origins and potential biases in the dataset, a brief data statement~\cite{datastatements} follows.
\begin{itemize}
    \item {\bf Curation rationale} Text was drawn from sources likely to hold debate and discussion, incorporating many different viewpoints. The dataset should be useful to those building systems to be applied to commentary on Russian news.
    \item {\bf Language variety} The BCP-47 descriptors relevant are ru-Cyrl-RU and ru-Cyrl-LV.
    \item {\bf Speaker demographic} Most speakers are L1 with other information hidden. It is assumed that all have internet access and most are adept internet users.
    \item {\bf Annotator demographic} Annotators are males between 25-40 of European descent, with high degrees of education.
    \item {\bf Speech situation} Utterances are from November 2017, written, without editing, and spontaneous, in a public internet conversation context.
\end{itemize}
%!TEX root = rustance.tex

\section{Implementation}
\label{chap:met}

As a baseline, and to provide a platform for analysis of the data, we built a pipeline of corpora preprocessing, feature extaction and classifier training. 

\subsection{Preprocessing}

Dealing with natural languages is often a complicated task with many caveats; this is no better with social media.
Phenomena prevalent in social media text include typos, acronyms, slang and another examples of non-dictionary and unexpected words.
This can mean that finding representations for words can be very noisy, because the embedding models are usually trained using normal and grammatically correct corpora.
In the case of RuStance, this is exacerbated by the paucity of large Russian datasets or embedding collections. 

In order to be able to process in a form of vectors in a meaningful n-dimensional space, the texts of the dataset have to be converted from a human friendly representation to \textit{Word Embeddings}~\cite{mikolov2013distributed}.
First, a strict filtering step removes all of the social-media-specific entities like \textit{\@-like} mentions, hashtags, URLs. 
We decided to proceed only with words and standard punctuation.

To convert a token to the expected format we had to pick a threshold that would cut out the outliers. % LD - what's going on here?
Also, by having a fixed length input (array of tokens) it is possible to fit smaller texts into the input by substituting the missing tokens with zero-tokens. 

\cite*{KutuzovKuzmenko2017} provide a pretrained model of word embeddings for Russian called RusVectores. 
Since our dataset covers mostly web and social conversations, we used a model that was trained on a web corpora and Wikipedia. 
This hopefully increases the probability that words from our dataset will be in the vocabulary.

\begin{figure*}
  \centering
  \includegraphics[width=\textwidth]{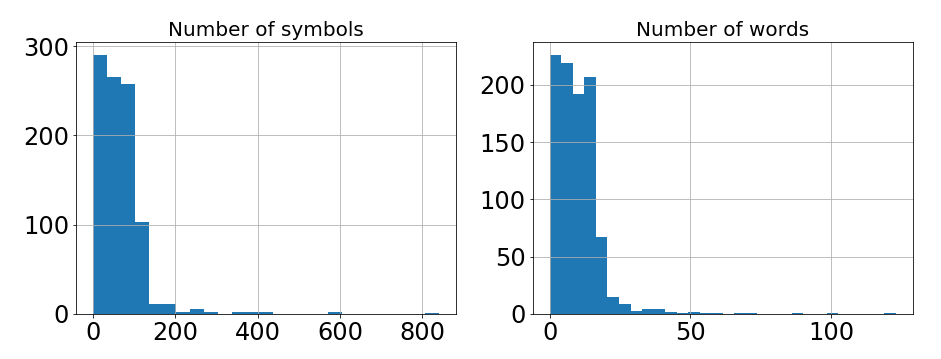}
  \caption{ Dataset symbols/words length distribution. }
  \label{fig:words_length_distribution}
\end{figure*}

To maintain a uniform feature representation, we set the input length to 25 words.
We set the length of the input to be 25 words because over 95\% of the records would pass the threshold (Figure~\ref{fig:words_length_distribution}) and the outliers will be abandoned. For the records that have less than 25 words we used pad 
sequence from Keras \cite{chollet2015keras} which appends empty word embeddings in order to equalize lengths. Unknown tokens (the tokens that are not in the vocabulary) are substituted with zero-vectors. 

So far an entry of the dataset for the classifier has 25 features = word embeddings and a class label. Then features can be used as an input for both Bayesian and Deep Learning models. At this step we have prepared dataset that later will be split for delayed cross-validation, model training and evaluation.

After the tokenization Keras Tokenizer is next to be trained on the texts to substitute words with Term Frequency - Inverse Document Frequency [TFIDF] indices. Using that indices and Gensim package~\cite{gensim} we create Embedding layer of the Deep Learning model~\cite{KutuzovKuzmenko2017}. 

It is known from the documentation that the models are initialized with the most common defaults, thus we expect our models to perform at a non-zero level.
%!TEX root = rustance.tex

\section{Evaluation and Discussion}

The dataset is split into train and test partitions, and then perform cross-validation.

%Cross-Validation allows to find hyperparamters over a grid by testing model's performance for each combination over a range parameters. 
%This procudure generates combinations of parameter, trains models with that parameters and saves the evaluation metrics afterwards. 
%When all of the parameters have been test it returns the best paramters according to the previously memorized value. 

We evaluate model performance using \textit{accuracy} and \textit{f1-measure}.
The dataset consists of short messages, so accuracy will be more representative in the context of overall analysis, whereas f1-measure will partially compensate the imbalance of classes. 
In this case, accuracy tends to be more effective, since we want to exclude as many false positives as possible in order to not to call arbitrary media to be fake. % LD check

% \begin{table}[h]
% \centering
% \caption{Evaluation metrics after Cross-validation, k = 5}
% \label{my-label}
% \begin{tabular}{c|ccccc}
%  & Bagging & AdaBoost & Boosting & SGD & Logistic Regression \\ \hline
% F1 & \textbf{0.832} & 0.530 & \textbf{0.865} & 0.266 & 0.259 \\
% Accuracy & \textbf{0.925} & 0.766 & \textbf{0.925} & 0.582 & 0.678
% \end{tabular}
% \end{table}

\begin{table*}[t]
  \centering
  \caption{Evaluation metrics after Cross-validation, k = 5}
  \label{fig:metrics}
  \begin{tabular}{c|ccccc}
   & Bagging & AdaBoost & Boosting & SGD & Logistic Regression \\ \hline
  F1 & \textbf{0.832} & 0.530 & \textbf{0.865} & 0.266 & 0.259 \\
  Accuracy & \textbf{0.925} & 0.766 & \textbf{0.925} & 0.582 & 0.678
  \end{tabular}
\end{table*}

\begin{figure*}
  \centering
  \includegraphics[width=\textwidth,scale=1.5]{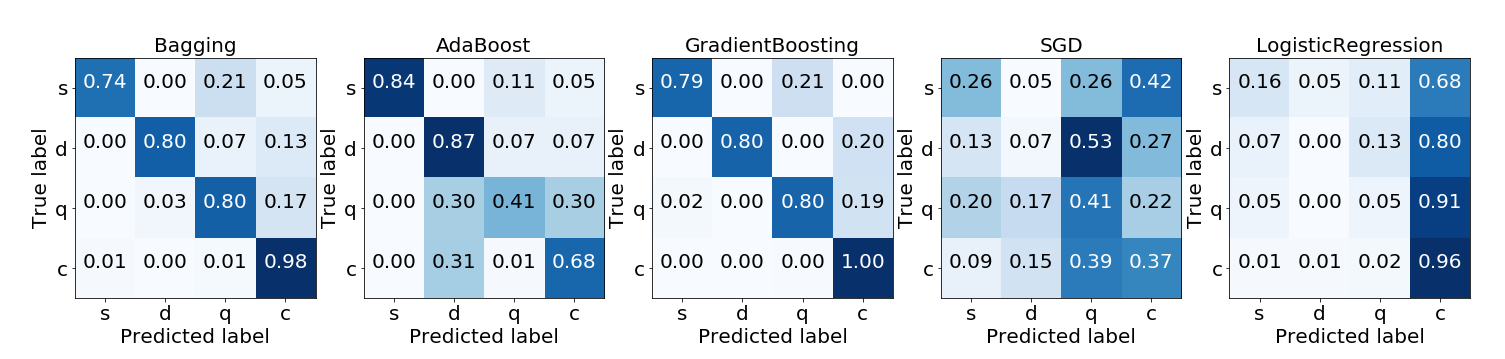}
  \caption{ Confusion matrices for SQDC classification of stance on RuStance. }
  \label{fig:confusion_matrices}
\end{figure*}

With trained split stratified validation and precision as a metric we built \textit{Confusion Matrices} (fig. \ref{fig:confusion_matrices}) for top-scoring classifiers. The fit-predict part was performed with \textit{GridSearch} cross-validation with K equals 5 and train/test split coefficient equals 0.1.

\subsection{Analysis}

To analyse misclassification errors, we generated confusion matrixes.
These enable visual comparison of accurate and error predictions for each class. 
The ideal outcome of such a visualization would be a unit matrix since the main diagonal would contain ones which means that every class was predicted 100\% correct. 
In practice, the stronger (close to 1) the main diagonal -- the better the classifier performs. 
%Also, it is an intuitive approach to see the cases where the model tends to overfit.

According to the confusion matrices one can infer that models tend to overfitting by  predicting \textit{Comment}-class all of the time, which would result in around 70\% of true positives. However, the models that use Bagging and Boosting are more reliable and predict all of the classes more or less equally.
%!TEX root = rustance.tex

\section{Related Work}
\label{chap:related}

Evaluation of the reliability of the information is a difficult and time consuming task, even for trained professionals \cite{medium}. Fortunately, the process can be divided into steps or stages, some of which can later be automated.

The first step towards classifying claims as either fake or trustworthy is to find out how others react to said claims. 
This process of Stance Detection plays significant role in fact-checking pipelines~\cite{mohammad2017stance,fnc1winner,medium}.

Since initial work on determining veracity of social media~\cite{qazvinian2011rumor}, powerful systems and annotation schemes have been developed to support the analysis of rumours and misinformation in text.
In this paper, the authors introduced methods to enhance the performance of classifiers trained on relatively long-term rumours in tweets. 
The idea was later extended by \cite{liu2015real} who, in addition, tracked the presence of positive or negative markers in a tweet.
Later \cite{ferreira2016emergent} provides a novel hand crafted dataset of rumoured claims. Our work is to provide a similar dataset for Russian tweets.

\cite{lukasik2016hawkes} exploited the temporal sequence of tweets, although the conversational structure was ignored and each tweet was treated as a separate unit. 
In other domains where debates or conversations are involved, the sequence of responses has been exploited to make the most of the evolving discourse and perform an improved classification of each individual post after learning the structure and dynamics of the conversation as a whole.

Classification of stance towards a claim on Twitter has been mentioned in SemEval-2016 task 6 \cite{mohammad2017stance}. 
Subtask B tested stance detection towards an unlabelled target, which required a weakly supervised or unsupervised approach. 
The dataset of this competition was not related to rumours or breaking news, it only considered a 3-way classification and did not provide any relations between tweets, which were treated as individual instances.

Recent state-of-the-art work on stance classification includes the top-scoring system in the RumourEval exercise~\cite{derczynski2017semeval}, which decomposed conversational branches into lines for use as context input to an LSTM~\cite{kochkina}, and another approach based on proximity to English words representative of certain stance classes~\cite{aker}.
Stance detection has also been used to develop datasets of diverse stance, enabling construction of balanced summaries and examination of arugmentation and counter-argumentation~\cite{ruder360}.
Finally, while RumourEval, the Fake News Challenge, FEVER~\cite{thornefever} and others have provided datasets, and some have developed creative uses of this stance data, all of these resources are in English; RuStance is the first such dataset for Russian, a language hosting an active, nuanced and passionate political debate.

%!TEX root = rustance.tex

\section{Conclusion}

\textit{RuStance} \cite{lozhnikov_derczynski_mazzara_2018} is the first dataset for stance classification in Russian. 
We think it is a highly area to study, collecting data for future researchers with tools and data, and opening up the arena of fake news in Russia to global researchers. 
It comprises a dataset from multiple sources, including many conversation threads, and a mixture of social interaction.

%We aiming to create a multilingual dataset that will enable process and embed most of the social media conversation because you might have noticed that non-native speakers tend to use English words and phrases which means that as of now it is unavailable for us to make word embeddings work with only Russian datasets.

A baseline is included.
As we assumed, a thousand examples is definitely not enough to fit the LSTM layer. 
Nevertheless, we achieved accuracy over 90\% using classifiers without any tuning whatsoever.
Confusion matrices suggest that even if a single precision rate is high the class imbalance is still a huge issue and the bottleneck that stops us from fitting a really accurate classifiers.
%Using regularization techniques one could try to extract the most important features and introduce additional heuristics above it -- this is 

%Future work requires dataset extension first of all. 
%Social Media are easy to grab and to search for claims. 
%Which also enables decent activity of the Fake News makers and spreaders.

%It is important to pay attention to the embeddings part and the Keras Neural Network construction. There might be tricks to embed texts in a more efficient way rather than TF-IDF index substitution. But the tricky part would than be to have a large dataset for embeddings training that has most of the social media conversations.

New metrics are also to be considered and developed. 
Russian tweets seem to be as reliable as English in terms of consistency and class distribution. 
%It is not mandatory to be a native speaker to label some texts, however it is highly suggested.
This dataset and baseline provide first steps into analyzing fake news spread and generation among Russian speakers, and we hope, with further work, multilingually.

\iffalse
\section{Acknowledgments*}
barbara - input
isabelle - input
innopolis - support
\fi

\bibliographystyle{ieeetr}
\bibliography{ms}

\begin{thebibliography}{10}

\bibitem{forbes}
K.~Rapoza, ``{These Two Russian 'Fake News' Outfits Get Billions Of Hits On
  Facebook}.''
  https://www.forbes.com/sites/kenrapoza/2017/09/22/these-two-russian-fake-news-outfits-get-billions-of-hits-on-facebook,
  2017.

\bibitem{2017arXiv170400656Z}
A.~{Zubiaga}, A.~{Aker}, K.~{Bontcheva}, M.~{Liakata}, and R.~{Procter},
  ``{Detection and Resolution of Rumours in Social Media: A Survey},'' {\em
  ArXiv e-prints}, Apr. 2017.

\bibitem{mrowcastance}
D.~Mrowca, E.~Wang, and A.~Kosson, ``Stance detection for fake news
  identification,'' 2017.

\bibitem{ferreira2016emergent}
W.~Ferreira and A.~Vlachos, ``Emergent: a novel data-set for stance
  classification,'' in {\em Proceedings of the 2016 Conference of the North
  American Chapter of the Association for Computational Linguistics: Human
  Language Technologies}, ACL, 2016.

\bibitem{anta2013sentiment}
A.~F. Anta, L.~N. Chiroque, P.~Morere, and A.~Santos, ``Sentiment analysis and
  topic detection of spanish tweets: A comparative study of of nlp
  techniques,'' {\em Procesamiento del lenguaje natural}, vol.~50, pp.~45--52,
  2013.

\bibitem{taule2017overview}
M.~Taul{\'e}, M.~A. Mart{\'i}, F.~M. Rangel, P.~Rosso, C.~Bosco, V.~Patti, {\em
  et~al.}, ``Overview of the task on stance and gender detection in tweets on
  catalan independence at ibereval 2017,'' in {\em 2nd Workshop on Evaluation
  of Human Language Technologies for Iberian Languages, IberEval 2017},
  vol.~1881, pp.~157--177, CEUR-WS, 2017.

\bibitem{enikolopov2011media}
R.~Enikolopov, M.~Petrova, and E.~Zhuravskaya, ``Media and political
  persuasion: Evidence from {Russia},'' {\em American Economic Review},
  vol.~101, no.~7, pp.~3253--85, 2011.

\bibitem{zubiaga2016stance}
A.~Zubiaga, E.~Kochkina, M.~Liakata, R.~Procter, and M.~Lukasik, ``Stance
  classification in rumours as a sequential task exploiting the tree structure
  of social media conversations,'' {\em arXiv preprint arXiv:1609.09028}, 2016.

\bibitem{qazvinian2011rumor}
V.~Qazvinian, E.~Rosengren, D.~R. Radev, and Q.~Mei, ``Rumor has it:
  Identifying misinformation in microblogs,'' in {\em Proceedings of the
  Conference on Empirical Methods in Natural Language Processing},
  pp.~1589--1599, Association for Computational Linguistics, 2011.

\bibitem{lukasik2016hawkes}
M.~Lukasik, P.~Srijith, D.~Vu, K.~Bontcheva, A.~Zubiaga, and T.~Cohn, ``Hawkes
  processes for continuous time sequence classification: an application to
  rumour stance classification in twitter,'' in {\em Proceedings of 54th Annual
  Meeting of the Association for Computational Linguistics}, pp.~393--398,
  Association for Computational Linguistics, 2016.

\bibitem{meduza}
{Meduza}, ``{Meduza.io}.'' \url{http://meduza.io}, 2018.

\bibitem{russiatoday}
{Channel RT TV}, ``{Russia Today}.'' \url{https://rt.com}, 2018.

\bibitem{guardian2017russia}
{The Guardian}, ``Russia's 'irrefutable evidence' of us help for isis appears
  to be video game still.''
  \url{https://www.theguardian.com/world/2017/nov/14/russia-us-isis-syria-video-game-still},
  2017.

\bibitem{meduza2017mod}
Meduza, ``{Meduza.io: On Fake Evidence}.''
  https://meduza.io/shapito/2017/11/14/minoborony-vylozhilo-neosporimoe-dokazatelstvo-sotrudnichestva-ssha-i-ig-skrinshot-iz-mobilnoy-igry,
  2017.

\bibitem{meduza2017mod2}
Meduza, ``{Meduza.io: On Ministry of Defense}.''
  https://meduza.io/shapito/2017/11/14/minoborony-vylozhilo-neosporimoe-dokazatelstvo-sotrudnichestva-ssha-i-ig-skrinshot-iz-mobilnoy-igry,
  2017.

\bibitem{rt2017}
{Channel RT TV}, ``{Russia Today: On Russian President Candidates 2018}.''
  https://russian.rt.com/inotv/2017-11-14/Rukovoditel-internet-kampanii-Sobchak-eyo-uchastie,
  2017.

\bibitem{derczynski2014pheme}
L.~Derczynski and K.~Bontcheva, ``Pheme: Veracity in digital social
  networks.,'' in {\em UMAP Workshops}, 2014.

\bibitem{fakenewschallenge}
D.~Pomerleau and D.~Rao, ``Fake news challenge.''
  http://www.fakenewschallenge.org, 2017.

\bibitem{datastatements}
Anonymous, ``{Data Statements for NLP: Toward Mitigating System Bias and
  Enabling Better Science}.'' OpenReview.net, 2018.

\bibitem{mikolov2013distributed}
T.~Mikolov, I.~Sutskever, K.~Chen, G.~S. Corrado, and J.~Dean, ``Distributed
  representations of words and phrases and their compositionality,'' in {\em
  Advances in neural information processing systems}, pp.~3111--3119, 2013.

\bibitem{lozhnikov_derczynski_mazzara_2018}
``Rustance.'' \url{https://figshare.com/articles/dataset_csv/7151906/2}.

\bibitem{pangopinion}
B.~Pang and L.~Lee, {\em Opinion mining and sentiment analysis}.
\newblock Now Publishers Inc., Foundations trends in information retrieval,
  available at http://portal. acm. org/citation. cfm, 2008.

\bibitem{liu2010sentiment}
B.~Liu, ``Sentiment analysis and subjectivity.,'' {\em Handbook of natural
  language processing}, vol.~2, pp.~627--666, 2010.

\bibitem{go2009twitter}
A.~Go, R.~Bhayani, and L.~Huang, ``Twitter sentiment classification using
  distant supervision,'' {\em CS224N Project Report, Stanford}, vol.~1, no.~12,
  2009.

\bibitem{laboreiro2010tokenizing}
G.~Laboreiro, L.~Sarmento, J.~Teixeira, and E.~Oliveira, ``Tokenizing
  micro-blogging messages using a text classification approach,'' in {\em
  Proceedings of the fourth workshop on Analytics for noisy unstructured text
  data}, pp.~81--88, ACM, 2010.

\bibitem{pak2010twitter}
A.~Pak and P.~Paroubek, ``Twitter as a corpus for sentiment analysis and
  opinion mining.,'' in {\em LREC}, 2010.

\bibitem{kukich1992techniques}
K.~Kukich, ``Techniques for automatically correcting words in text,'' {\em Acm
  Computing Surveys (CSUR)}, vol.~24, no.~4, pp.~377--439, 1992.

\bibitem{bermingham2010classifying}
A.~Bermingham and A.~F. Smeaton, ``Classifying sentiment in microblogs: is
  brevity an advantage?,'' in {\em Proceedings of the 19th ACM international
  conference on Information and knowledge management}, pp.~1833--1836, ACM,
  2010.

\bibitem{engstrom2004topic}
C.~Engstrom, ``Topic dependence in sentiment classification,'' {\em Master's
  thesis, University of Cambridge}, 2004.

\bibitem{read2005using}
J.~Read, ``Using emoticons to reduce dependency in machine learning techniques
  for sentiment classification,'' in {\em Proceedings of the ACL student
  research workshop}, pp.~43--48, Association for Computational Linguistics,
  2005.

\bibitem{padro2010semantic}
L.~Padr{\'o}, S.~Reese, E.~Agirre, and A.~Soroa, ``Semantic services in
  freeling 2.1: Wordnet and ukb,'' in {\em 5th Global WordNet Conference},
  pp.~99--105, 2010.

\bibitem{chang2011libsvm}
C.-C. Chang and C.-J. Lin, ``Libsvm: a library for support vector machines,''
  {\em ACM transactions on intelligent systems and technology (TIST)}, vol.~2,
  no.~3, p.~27, 2011.

\bibitem{mathioudakis2010twittermonitor}
M.~Mathioudakis and N.~Koudas, ``Twittermonitor: trend detection over the
  twitter stream,'' in {\em Proceedings of the 2010 ACM SIGMOD International
  Conference on Management of data}, pp.~1155--1158, ACM, 2010.

\bibitem{vakali2012social}
A.~Vakali, M.~Giatsoglou, and S.~Antaris, ``Social networking trends and
  dynamics detection via a cloud-based framework design,'' in {\em Proceedings
  of the 21st International Conference on World Wide Web}, pp.~1213--1220, ACM,
  2012.

\bibitem{sklearn_api}
L.~Buitinck, G.~Louppe, M.~Blondel, F.~Pedregosa, A.~Mueller, O.~Grisel,
  V.~Niculae, P.~Prettenhofer, A.~Gramfort, J.~Grobler, R.~Layton,
  J.~VanderPlas, A.~Joly, B.~Holt, and G.~Varoquaux, ``{API} design for machine
  learning software: experiences from the scikit-learn project,'' in {\em ECML
  PKDD Workshop: Languages for Data Mining and Machine Learning}, pp.~108--122,
  2013.

\bibitem{chollet2015keras}
F.~Chollet {\em et~al.}, ``Keras.'' \url{https://github.com/keras-team/keras},
  2015.

\bibitem{hermann2015teaching}
K.~M. Hermann, T.~Kocisky, E.~Grefenstette, L.~Espeholt, W.~Kay, M.~Suleyman,
  and P.~Blunsom, ``Teaching machines to read and comprehend,'' in {\em
  Advances in Neural Information Processing Systems}, pp.~1693--1701, 2015.

\bibitem{wang2017bilateral}
Z.~Wang, W.~Hamza, and R.~Florian, ``Bilateral multi-perspective matching for
  natural language sentences,'' {\em arXiv preprint arXiv:1702.03814}, 2017.

\bibitem{augenstein2016stance}
I.~Augenstein, T.~Rockt{\"a}schel, A.~Vlachos, and K.~Bontcheva, ``Stance
  detection with bidirectional conditional encoding,'' {\em arXiv preprint
  arXiv:1606.05464}, 2016.

\bibitem{wang2011detection}
W.~Wang, S.~Yaman, K.~Precoda, C.~Richey, and G.~Raymond, ``Detection of
  agreement and disagreement in broadcast conversations,'' in {\em Proceedings
  of the 49th Annual Meeting of the Association for Computational Linguistics:
  Human Language Technologies: short papers-Volume 2}, pp.~374--378,
  Association for Computational Linguistics, 2011.

\bibitem{zubiaga2015towards_}
A.~Zubiaga, M.~Liakata, R.~Procter, K.~Bontcheva, and P.~Tolmie, ``Towards
  detecting rumours in social media,'' in {\em AAAI Workshop: AI for Cities},
  2015.

\bibitem{abbott2011can}
R.~Abbott, M.~Walker, P.~Anand, J.~E. Fox~Tree, R.~Bowmani, and J.~King, ``How
  can you say such things?!?: Recognizing disagreement in informal political
  argument,'' in {\em Proceedings of the Workshop on Languages in Social
  Media}, pp.~2--11, Association for Computational Linguistics, 2011.

\bibitem{fitzgerald2011exploiting}
N.~FitzGerald, G.~Carenini, G.~Murray, and S.~Joty, ``Exploiting conversational
  features to detect high-quality blog comments,'' {\em Advances in Artificial
  Intelligence}, pp.~122--127, 2011.

\bibitem{nakov2015semeval}
P.~Nakov, L.~M{\`a}rquez, W.~Magdy, A.~Moschitti, J.~Glass, and B.~Randeree,
  ``Semeval-2015 task 3: Answer selection in community question answering.,''
  {\em SemEval@ NAACL-HLT}, vol.~2015, 2015.

\bibitem{qu2011interactive}
Z.~Qu and Y.~Liu, ``Interactive group suggesting for twitter,'' in {\em
  Proceedings of the 49th Annual Meeting of the Association for Computational
  Linguistics: Human Language Technologies: short papers-Volume 2},
  pp.~519--523, Association for Computational Linguistics, 2011.

\bibitem{zhao2015enquiring}
Z.~Zhao, P.~Resnick, and Q.~Mei, ``Enquiring minds: Early detection of rumors
  in social media from enquiry posts,'' in {\em Proceedings of the 24th
  International Conference on World Wide Web}, pp.~1395--1405, International
  World Wide Web Conferences Steering Committee, 2015.

\bibitem{liu2015real}
X.~Liu, A.~Nourbakhsh, Q.~Li, R.~Fang, and S.~Shah, ``Real-time rumor debunking
  on twitter,'' in {\em Proceedings of the 24th ACM International on Conference
  on Information and Knowledge Management}, pp.~1867--1870, ACM, 2015.

\bibitem{enayet2017niletmrg}
O.~Enayet and S.~R. El-Beltagy, ``Niletmrg at semeval-2017 task 8: Determining
  rumour and veracity support for rumours on twitter.,'' in {\em Proceedings of
  the 11th International Workshop on Semantic Evaluation (SemEval-2017)},
  pp.~470--474, 2017.

\bibitem{bahuleyan2017uwaterloo}
H.~Bahuleyan and O.~Vechtomova, ``Uwaterloo at semeval-2017 task 8: Detecting
  stance towards rumours with topic independent features,'' in {\em Proceedings
  of the 11th International Workshop on Semantic Evaluation (SemEval-2017)},
  pp.~461--464, 2017.

\bibitem{zubiaga2016analysing}
A.~Zubiaga, M.~Liakata, R.~Procter, G.~W.~S. Hoi, and P.~Tolmie, ``Analysing
  how people orient to and spread rumours in social media by looking at
  conversational threads,'' {\em PloS one}, vol.~11, no.~3, p.~e0150989, 2016.

\bibitem{shao2016hoaxy}
C.~Shao, G.~L. Ciampaglia, A.~Flammini, and F.~Menczer, ``Hoaxy: A platform for
  tracking online misinformation,'' in {\em Proceedings of the 25th
  International Conference Companion on World Wide Web}, pp.~745--750,
  International World Wide Web Conferences Steering Committee, 2016.

\bibitem{zhang2015automatic}
Q.~Zhang, S.~Zhang, J.~Dong, J.~Xiong, and X.~Cheng, ``Automatic detection of
  rumor on social network,'' in {\em Natural Language Processing and Chinese
  Computing}, pp.~113--122, Springer, 2015.

\bibitem{kumar2014detecting}
K.~K. Kumar and G.~Geethakumari, ``Detecting misinformation in online social
  networks using cognitive psychology,'' {\em Human-centric Computing and
  Information Sciences}, vol.~4, no.~1, p.~14, 2014.

\bibitem{KutuzovKuzmenko2017}
A.~Kutuzov and E.~Kuzmenko, {\em {WebVectors: A Toolkit for Building Web
  Interfaces for Vector Semantic Models}}, pp.~155--161.
\newblock Cham: Springer International Publishing, 2017.

\bibitem{goldberg2014word2vec}
Y.~Goldberg and O.~Levy, ``word2vec explained: deriving mikolov et al.'s
  negative-sampling word-embedding method,'' {\em arXiv preprint
  arXiv:1402.3722}, 2014.

\bibitem{nytimes}
S.~Tavernise, ``{As Fake News Spreads Lies, More Readers Shrug at the Truth}.''
  https://www.nytimes.com/2016/12/06/us/fake-news-partisan-republican-democrat.html,
  2016.

\bibitem{journalims}
J.~H. Michael~Barthel, Amy~Mitchell, ``Many americans believe fake news is
  sowing confusion.''
  http://www.journalism.org/2016/12/15/many-americans-believe-fake-news-is-sowing-confusion,
  2016.

\bibitem{medium}
D.~Ghulati, ``Introducing factmata — artificial intelligence for political
  fact-checking.''
  https://medium.com/factmata/introducing-factmata-artificial-intelligence-for-political-fact-checking-db8acdbf4cf1,
  2016.

\bibitem{ruder2017}
S.~Ruder, ``On word embeddings.'' \url{http://ruder.io/word-embeddings-1},
  2016.

\bibitem{fnc1winner}
Y.~P. Sean~Baird, Doug~Sibley, ``Talos targets disinformation with fake news
  challenge victory.''
  \url{http://blog.talosintelligence.com/2017/06/talos-fake-news-challenge.html},
  2017.

\bibitem{fnc1second}
A.~Hanselowski, ``Team athene on the fake news challenge.''
  \url{https://medium.com/@andre134679/team-athene-on-the-fake-news-challenge-28a5cf5e017b},
  2017.

\bibitem{gensim}
R.~{\v R}eh{\r u}{\v r}ek and P.~Sojka, ``{Software Framework for Topic
  Modelling with Large Corpora},'' in {\em {Proceedings of the LREC 2010
  Workshop on New Challenges for NLP Frameworks}}, (Valletta, Malta),
  pp.~45--50, ELRA, May 2010.
\newblock \url{http://is.muni.cz/publication/884893/en}.

\bibitem{riedel2017simple}
B.~Riedel, I.~Augenstein, G.~P. Spithourakis, and S.~Riedel, ``A simple but
  tough-to-beat baseline for the fake news challenge stance detection task,''
  {\em arXiv preprint arXiv:1707.03264}, 2017.

\bibitem{rakholiatrue}
N.~Rakholia and S.~Bhargava, ``“is it true?”--deep learning for stance
  detection in,''

\bibitem{krejzl2017stance}
P.~Krejzl, B.~Hourov{\'a}, and J.~Steinberger, ``Stance detection in online
  discussions,'' {\em arXiv preprint arXiv:1701.00504}, 2017.

\bibitem{derczynski2017semeval}
L.~Derczynski, K.~Bontcheva, M.~Liakata, R.~Procter, G.~W.~S. Hoi, and
  A.~Zubiaga, ``Semeval-2017 task 8: Rumoureval: Determining rumour veracity
  and support for rumours,'' {\em arXiv preprint arXiv:1704.05972}, 2017.

\bibitem{mohammad2017stance}
S.~M. Mohammad, P.~Sobhani, and S.~Kiritchenko, ``Stance and sentiment in
  tweets,'' {\em ACM Transactions on Internet Technology (TOIT)}, vol.~17,
  no.~3, p.~26, 2017.

\bibitem{hamidian2016rumor}
S.~Hamidian and M.~T. Diab, ``Rumor identification and belief investigation on
  twitter.,'' in {\em Proceedings of WASSA at NAACL-HLT}, pp.~3--8, 2016.

\bibitem{aker}
A.~Aker, L.~Derczynski, and K.~Bontcheva, ``Simple open stance classification
  for rumour analysis,'' in {\em Proc. RANLP}, 2017.

\bibitem{ruder360}
S.~Ruder, J.~Glover, A.~Mehrabani, and P.~Ghaffari, ``360${}^\circ$ stance
  detection,'' in {\em Proceedings of the 2018 Conference of the North American
  Chapter of the Association for Computational Linguistics: Demonstrations},
  pp.~31--35, Association for Computational Linguistics, 2018.

\bibitem{thornefever}
J.~Thorne, A.~Vlachos, C.~Christodoulopoulos, and A.~Mittal, ``Fever: a
  large-scale dataset for fact extraction and verification,'' in {\em
  Proceedings of the 2018 Conference of the North American Chapter of the
  Association for Computational Linguistics: Human Language Technologies,
  Volume 1 (Long Papers)}, pp.~809--819, Association for Computational
  Linguistics, 2018.

\bibitem{kochkina}
E.~Kochkina, M.~Liakata, and I.~Augenstein, ``{Turing at SemEval-2017 Task 8:
  Sequential Approach to Rumour Stance Classification with Branch-LSTM},'' in
  {\em Proc. SemEval}, 2017.

\end{thebibliography}

\end{document}